\documentclass[runningheads]{llncs}

\usepackage[mobile]{eccv}

\usepackage{eccvabbrv}
\usepackage{xcolor}
\usepackage{xcolor,colortbl}
\usepackage{color}
\usepackage{colortbl}
\usepackage{bm}
\usepackage{multicol}
\usepackage{enumitem}
\usepackage{wrapfig}
\usepackage{graphicx}
\usepackage{booktabs}

\usepackage[accsupp]{axessibility}  

\definecolor{amber}{rgb}{1.0, 0.75, 0.0}

\usepackage{hyperref}

\usepackage{orcidlink}

\begin{document}

\title{Hash3D: Training-free Acceleration for \\3D Generation} 
\titlerunning{Hash3D}

\author{Xingyi Yang \quad Xinchao Wang}

\institute{National University of Singapore \\
\email{xyang@u.nus.edu}, \email{xinchao@nus.edu.sg}}

\authorrunning{X. Yang et al.}


\maketitle
\vspace{-2mm} 
\begin{center}
    \centering
    \captionsetup{type=figure}
\includegraphics[width=\textwidth]{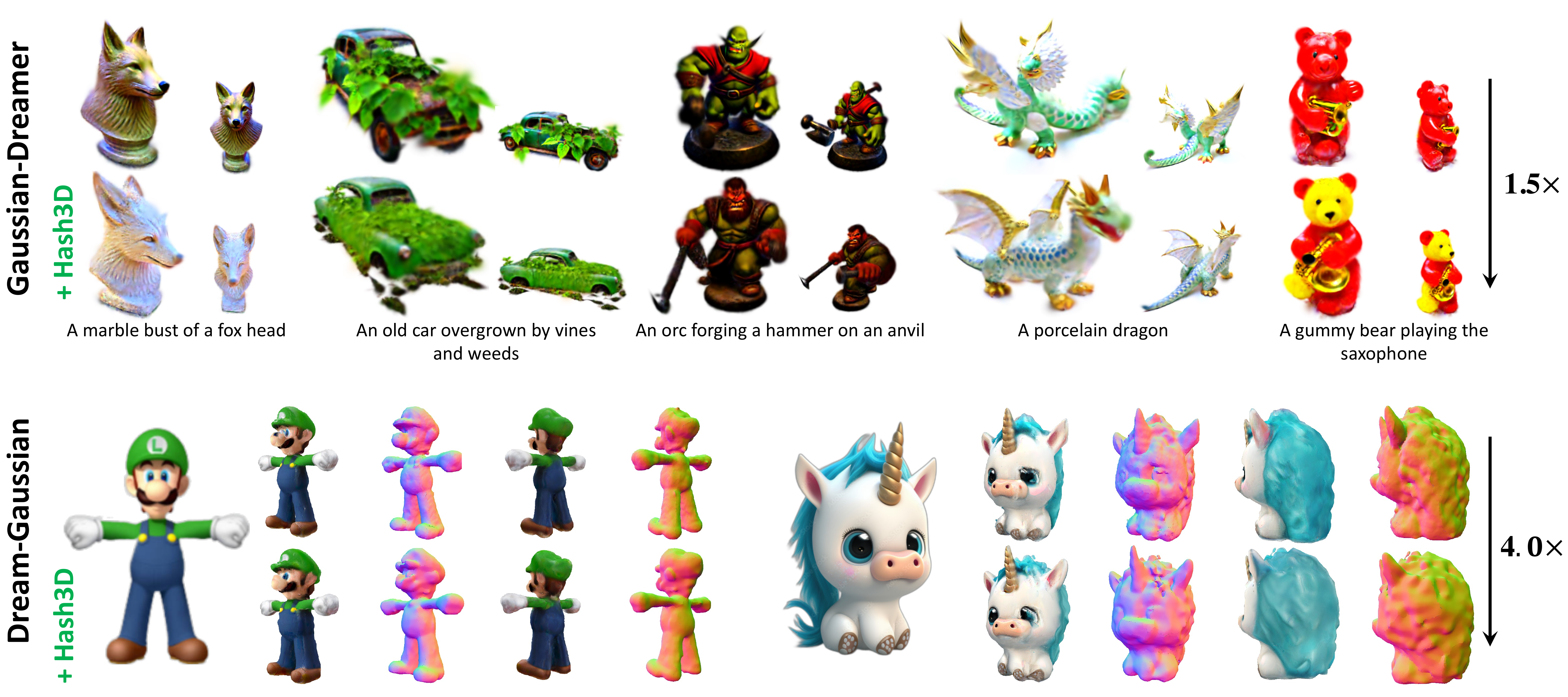}
    \vspace{-6mm}
    \caption{Examples by applying our Hash3D on Gaussian-Dreamer~\cite{yi2023gaussiandreamer} and Dream-Gaussian~\cite{tang2023dreamgaussian}. We accelerate Gaussian-Dreamer by $1.5\times$ and Dream-Gaussian by $4\times$ with comparable visual quality.
    }
    \label{fig:head}
\end{center}%

\begin{abstract}
The evolution of 3D generative modeling has been notably propelled by the adoption of 2D diffusion models. Despite this progress, 
the cumbersome optimization process per se
presents a critical hurdle to efficiency. 
In this paper,
we introduce Hash3D, 
a universal acceleration for 3D generation without model training.
Central to Hash3D is the insight that feature-map redundancy is prevalent in images rendered from camera positions and diffusion time-steps in close proximity. By effectively hashing and reusing these feature maps across neighboring timesteps and camera angles, Hash3D substantially prevents redundant calculations, thus accelerating the diffusion model's inference in 3D generation tasks. We achieve this through an adaptive grid-based hashing.
Surprisingly, this feature-sharing mechanism 
not only speed up the generation 
but also enhances the smoothness and view consistency of the synthesized 3D objects. Our experiments covering 5 text-to-3D and 3 image-to-3D models, demonstrate Hash3D’s versatility to speed up optimization, enhancing efficiency by $1.3\sim 4\times$. Additionally, Hash3D's integration with 3D Gaussian splatting largely speeds up 3D model creation, reducing text-to-3D processing to about 10 minutes and image-to-3D conversion to roughly 30 seconds. The project page is at~\url{https://adamdad.github.io/hash3D/}.
  \keywords{Fast 3D Generation \and Score Distillation Sampling}

\end{abstract}    
\section{Introduction}
\label{sec:intro}


In the evolving landscape of 3D generative modeling, the integration of 2D diffusion models~\cite{poole2023dreamfusion, wang2023score} has led to notable advancements. These methods leverage off-the-the-shelf image diffusion models to distill 3D models by predicting 2D score functions at different views, known as score distillation sampling~(SDS). 

While this approach has opened up new avenues for creating detailed 3D assets, it also brings forth significant challenges, particularly in terms of efficiency. Particularly, SDS requires sampling thousands of score predictions at different camera poses and denoising timesteps from the diffusion model, causing a extensively long optimization, even for hours to create one object~\cite{wang2024prolificdreamer}. These prolonged duration create a significant obstacle to apply them in practical application products, calling for new solutions to improve its efficiency.


To mitigate this bottleneck, current efforts concentrate on three strategies. The first strategy trains an inference-only models~\cite{instant3d2023,chen2023et3d,jun2023shape,xu2024dmvd,liu2024one} to bypass the lengthy optimization process. While effective, this method requires extensive training time and substantial computational resources. The second approach~\cite{tang2023dreamgaussian,yi2023gaussiandreamer,ren2023dreamgaussian4d} seeks to reduce optimization times through enhanced 3D parameterization techniques. However, this strategy necessitates a unique design for each specific representation, presenting its own set of challenges. The third approach attempts to directly generate sparse views to model 3D objects, assuming near-perfect view consistency in generation~\cite{kong2024eschernet,liu2024syncdreamer} which, in practice, is often not achievable.

Returning to the core issue within SDS, a considerable portion of computational effort is consumed in repeated sampling of the 2D image score function~\cite{song2019generative}. Motivated by methods that accelerate 2D diffusion sampling~\cite{song2021denoising,bao2022analyticdpm,lu2022dpm}, we posed the question: \emph{Is it possible to reduce the number of inference steps of the diffusion model for 3D generation?}

In pursuit of this, our exploration revealed a crucial observation: denoising outputs and feature maps from near camera positions and timesteps are remarkably similar. This discovery directly informs our solution, Hash3D, designed to reduce the computation by leveraging this redundancy.

At its core, Hash3D implements a space-time trade-off through a grid-based hash table. This table stores intermediate features from the diffusion model. Whenever a new sampled view is close to one it has already worked on, Hash3D efficiently retrieves the relevant features from the hash table. By reusing these features to calculate the current view’s score function, it avoids redoing calculations that have already been done. Additionally, we have developed a method to dynamically choose the grid size for each view, enhancing the system's adaptability. As such, Hash3D not only conserves computational resources, but does so without any model training or complex modifications, making it simple to implement and efficient to apply.

Beyond just being efficient, Hash3D helps produce 3D objects with improved multi-view consistency. Traditional diffusion-based methods often result in 3D objects with disjointed appearances when viewed from various angles~\cite{armandpour2023re}. In contrast, Hash3D connects independently sampled views by sharing features within each grid, leading to smoother, more consistent 3D models.

Another key advantage of Hash3D is on its versatility. It integrates seamlessly into a diverse array of diffusion-based 3D generative workflows. Our experiments, covering 5 text-to-3D and 3 image-to-3D models, demonstrate Hash3D's versatility to speed up optimization, enhancing efficiency by $1.3 \sim 4 \times$, without compromising on performance. Specifically, the integration of Hash3D with 3D Gaussian Splatting~\cite{kerbl3Dgaussians} brings a significant leap forward, cutting down the time for text-to-3D to about 10 minutes and image-to-3D  to roughly 30 seconds.

The contribution of this paper can be summarized into
\begin{itemize}
    \item We introduce the Hash3D, a versatile, plug-and-play and training-free acceleration method for diffusion-based text-to-3D and image-to-3D models. 
    \item The paper emphasizes the redundancy in diffusion models when processing nearby views and timesteps. This finding motivates the development of Hash3D, aiming to boost efficiency without compromising quality.
    \item Hash3D employs an adaptive grid-based hashing to efficiently retrieve features, significantly reducing the computations across view and time.  
    \item Our extensive testing across a range of models demonstrates that Hash3D not only speeds up the generative process by $1.3 \sim 4\times$, but also results in a slight improvement in performance.
\end{itemize}

\section{Related Work}

\noindent\textbf{3D Generation Model.} 
The development of 3D generative models has become a focal point in the computer vision. Typically, these models are trained to produce the parameters that define 3D representations. This approach has been successfully applied across several larger-scale models using extensive and diverse datasets for generating voxel representation~\cite{wu2016learning}, point cloud~\cite{achlioptas2018learning,nichol2022point}, implicit function~\cite{jun2023shap}, triplane~\cite{shue20233d,xu2024dmvd}. Despite these advances, scalability continues to be a formidable challenge, primarily due to data volume and computational resource constraints.
A promising solution to this issue lies in leveraging 2D generative models to enhance and optimize 3D representations. Recently, diffusion-based models, particularly those involving score distillation into 3D representations~\cite{poole2023dreamfusion}, represent significant progress. However, these methods are often constrained by lengthy optimization processes.

\noindent\textbf{Efficient Diffusion Model.} Diffusion models, known for their iterative denoising process for image generation, are pivotal yet time-intensive. There has been a substantial body of work aimed at accelerating these models. This acceleration can be approached from two angles: firstly, by reducing the sampling steps through advanced sampling mechanisms~\cite{song2021denoising, bao2022analyticdpm,liu2022pseudo,lu2022dpm} or timestep distillation~\cite{salimans2022progressive,song2023consistency}, which decreases the number of required sampling steps. The second approach focuses on minimizing the computational demands of each model inference. This can be achieved by developing smaller diffusion models~\cite{kim2023architectural,yang2023diffusion,fang2023structural} or reusing features from adjacent steps~\cite{ma2023deepcache,li2023faster}, thereby enhancing efficiency without compromising effectiveness. However, the application of these techniques to 3D generative tasks remains largely unexplored.


\noindent\textbf{Hashing Techniques.} Hashing, pivotal in computational and storage efficiency, involves converting variable-sized inputs into fixed-size hash code via \emph{hash functions}. These code index a \emph{hash table}, enabling fast and consistent data access. Widely used in file systems, hashing has proven effective in a variety of applications, like 3D representation~\cite{10.1145/2508363.2508374,muller2022instant,girish2023shacira,xie2023hollownerf}, neural network compression~\cite{10.5555/3045118.3045361,Kitaev2020Reformer}, using hashing as a components in deep network~\cite{roller2021hash} and neural network-based hash function development~\cite{7298947,zhu2016deep,cao2017hashnet,li2017deep}. Our study explores the application of hashing to retrieve features from 3D generation. By adopting this technique, we aim to reduce computational overhead for repeated diffusion sampling and speed up the creation of realistic 3D objects.


\section{Preliminary}

In this section, we provide the necessary notations, as well as the background on optimization-based 3D generation, focusing on diffusion models and Score Distillation Sampling (SDS)~\cite{poole2023dreamfusion}.

\subsection{Diffusion Models}

Diffusion models, a class of generative models, reverse a process of adding noise by constructing a series of latent variables. Starting with a dataset \(\mathbf{x}_0\) drawn from a distribution \(q(\mathbf{x}_0)\), the models progressively introduce Gaussian noise over \(T\) steps. Each step, defined as \(q(\mathbf{x}_t|\mathbf{x}_{t-1}) = \mathcal{N}(\mathbf{x}_{t};\sqrt{1-\beta_{t}}\mathbf{x}_{t-1}, \beta_{t}\mathbf{I})\), is controlled by \(\beta_{1:T}\), values ranging from 0 to 1. The inherently Gaussian nature of this noise enables direct sampling from \(q(\mathbf{x}_t)\) using the formula $\mathbf{x}_t = \sqrt{\bar{\alpha}_t} \mathbf{x}_0 + \sqrt{1-\bar{\alpha}_t} \bm{\epsilon}, \quad \text{where} \quad \epsilon \sim \mathcal{N}(0, \mathbf{I})$ with \(\alpha_t = 1-\beta_t\) and \(\bar{\alpha}_t = \prod_{s=1}^t \alpha_s\).

The reverse process is formulated as a variational Markov chain, parameterized by a time-conditioned denoising neural network \(\bm \epsilon(\mathbf{x}_t, t, y)\), with $y$ being the conditional input for generation, such as text for text-to-image model~\cite{saharia2022photorealistic} or camera pose for novel view synthesis~\cite{liu2023zero}.
The training of the denoiser aims to minimize a re-weighted evidence lower bound (ELBO), aligning with the noise:
\begin{align}
    \mathcal{L}_{\text{DDPM}} = \mathbb{E}_{t,\mathbf{x}_0, \bm{\epsilon}} \left[||\bm{\epsilon} - \bm \epsilon(\mathbf{x}_t, t, y) ||_2^2\right]
\end{align}
Here, \(\bm \epsilon(\mathbf{x}_t, t, y)\) approximates the score function \(\nabla_{\mathbf{x}_t} \log p(\mathbf{x}_t|\mathbf{x}_0)\). Data generation is achieved by denoising from noise, often enhanced using classifier-free guidance with scale parameter $\omega$: $\bm \hat{\epsilon}(\mathbf{x}_t, t, y) = (1+\omega) \bm \epsilon(\mathbf{x}_t, t, y) - \omega \bm \epsilon(\mathbf{x}_t, t, \emptyset)$.

\noindent\textbf{Extracting Feature from Diffusion Model.} A diffusion denoiser \(\bm \epsilon\) is typically parameterized with a U-Net~\cite{ronneberger2015u}. It uses \(l\) down-sampling layers  $\{D_i\}_{i=1}^l$ and up-sampling layers $\{U_i\}_{i=1}^l$, coupled with skip connections that link features from \(D_i\) to \(U_i\). This module effectively merges high-level features from \(U_{i+1}\) with low-level features from \(D_i\), as expressed by the equation:
\begin{equation}
    \mathbf{v}_{i+1}^{(U)} = \text{concat}(D_i(\mathbf{v}_{i-1}^{(D)}), U_{i+1}(\mathbf{v}_{i}^{(U)}))
\end{equation}
In this context, $\mathbf{v}_{i}^{(U)}$ and $\mathbf{v}_{i+1}^{(D)}$ represent the up-sampled and down-sampled features after the $i$-th layer, respectively.

\subsection{Score Distillation Sampling (SDS)}

The Score Distillation Sampling (SDS)~\cite{poole2023dreamfusion} represents an optimization-based 3D generation method. This method focuses on optimizing the 3D representation, denoted as $\Theta$, using a pre-trained 2D diffusion models with its noise prediction network, denoted as \(\bm{\epsilon}_{\text{pretrain}}(x_t, t, y)\).

Given a camera pose \(\bm c=(\theta,\phi,\rho)\in \mathbb{R}^3\) defined by elevation $\phi$, azimuth $\theta$ and camera distances $\rho$,  and the its corresponding prompt $y^c$, a differentiable rendering function \(g(\cdot;\Theta)\), SDS aims to refine the parameter \(\Theta\), such that each rendered image \(\bm{x}_0 = g(\bm c; \theta)\) is perceived as realistic by $\bm{\epsilon}_{\text{pretrain}}$. The optimization objective is formulated as follows:
\begin{equation}
    \min_{\Theta} \mathcal{L}_{\text{SDS}} = \mathbb{E}_{t,\bm c} \left[ \frac{\sigma_t}{\alpha_t} \omega(t) \text{KL}\left(q^\Theta(\bm{x}_t|y_c, t) \, \Vert \, p(\bm{x}_t|y_c; t)\right) \right] \label{eq:sds_full}
\end{equation}
By excluding the Jacobian term of the U-Net, the gradient of the optimization problem can be effectively approximated:
\begin{equation}
    \nabla_\Theta \mathcal{L}_{\text{SDS}} \approx \mathbb{E}_{t,\bm c,\bm{\epsilon}}\left[ \omega(t) (\bm{\epsilon}_{\text{pretrain}}(\bm{x}_t, t, y^c) - \bm{\epsilon})\frac{\partial \bm{x}}{\partial \Theta} \right]
    \label{eq:sds}
\end{equation}
To optimize Eq. \ref{eq:sds}, we randomly sample different time-step \(t\), camera \(\bm c\), and random noise \(\bm{\epsilon}\), and
compute gradient of the 3D representation, and update \(\theta\) accordingly. This approach ensures that the rendered image from 3D object aligns with the distribution learned by the diffusion model. 

\noindent\textbf{Efficiency Problem.} The main challenge lies in the need for thousands to tens of thousands of iterations to optimize Eq~\ref{eq:sds}, each requiring a separate diffusion model inference. This process is time-consuming due to the model's complexity. We make it faster by using a hash function to reuse features from similar inputs, cutting down on the number of calculations needed.

\section{Hash3D}

This section introduces Hash3D, a plug-and-play booster for Score Distillation Sampling~(SDS) to improve its efficiency. We start by analyzing the redundancy presented in the diffusion model across different timestep and camera poses. Based on the finding, we present our strategy that employ a grid-based hashing to reuse feature across different sampling iterations.

\subsection{Probing the Redundancy in SDS}
\label{sec:redundency}
\begin{figure*}[tb]
    \centering
    \includegraphics[width=\linewidth]{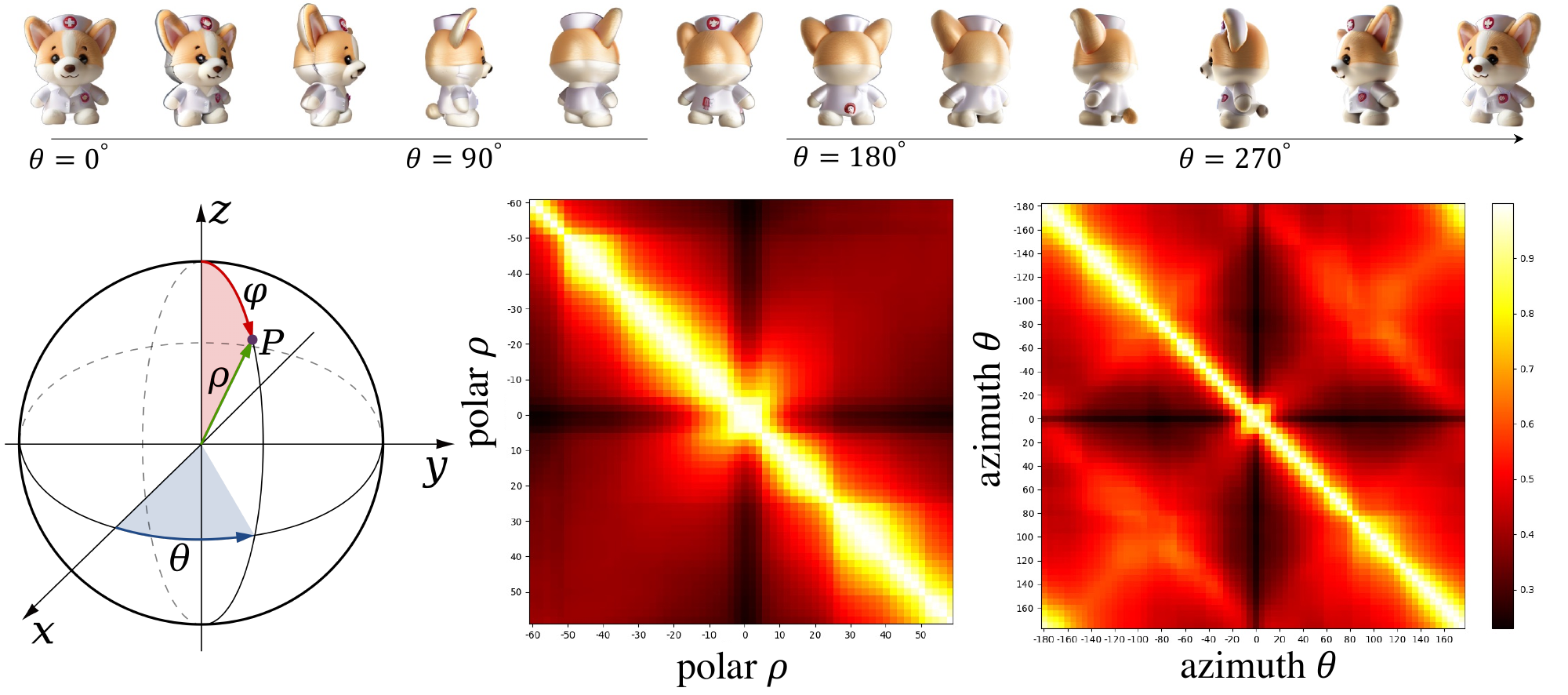}
     \vspace{-3mm}
    \caption{Feature similarity extracted from different camera poses. }
    \vspace{-5mm}
    \label{fig:simiarlity}
\end{figure*}


Typically, SDS randomly samples camera poses and timesteps to ensure that the rendered views align with diffusion model's prediction. A critical observation here is that deep feature extraction at proximate $\bm c$ and $t$ often reveals a high degree of similarity. This similarity underpins our method, suggesting that reusing features from nearby points does not significantly impact model's prediction.

\noindent\textbf{Measuring the Similarity.} Intuitively, images captured with close up camera and times results in similar visual information. We hypothesize that features produced by diffusion model would exhibit a similar pattern. In terms of the \emph{temporal similarity}, previous studies\cite{ma2023deepcache, li2023faster} have noted that features extracted from adjacent timesteps from diffusion models show a high level of similarity.

To test the hypothesis about the \emph{spatial similarity}, we conducted a preliminary study using the diffusion model to generate novel views of the same object from different camera positions. In practice, we use Zero-123~\cite{liu2023zero} to generate image from different cameras poses conditioned on single image input. For each specific camera angle and timestep, we extracted features $\mathbf{v}_{l-1}^{(U)}$ as the input of the last up-sampling layer at each timestep. By adjusting elevation angles ($\phi$) and azimuth angles ($\theta$), we were able to measure the cosine similarity of these features between different views, averaging the results across all timesteps. 

\begin{figure}[t]
    \centering
    \includegraphics[width=\linewidth]{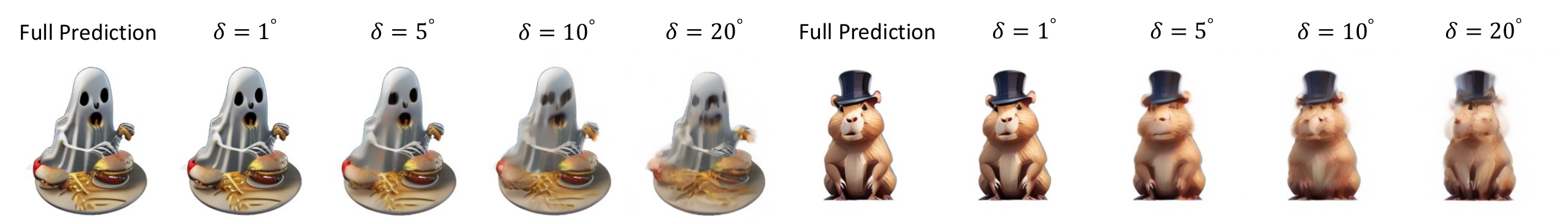}
    \vspace{-5mm}
    \caption{By interpolating latent between generated views, we enable the synthesis of novel views with no computations.}
    \label{fig:free-nv}
    \vspace{-5mm}
\end{figure}

The findings, presented in Figure~\ref{fig:simiarlity}, reveal a large similarity score in features from views within a $[-10^{\circ}, 10^{\circ}]$ range, with the value higher than 0.8. This phenomenon was not unique to Zero-123; we observed similar patterns in text-to-image diffusion models like Stable Diffusion~\cite{latentdiffusion}. These findings underscore the redundancy in predicted outputs within the SDS process.

\noindent\textbf{Synthesising Novel View for Free.} Exploiting redundancy, we conducted an initial experiment to create new views by simply reusing and interpolating scores from precomputed nearby cameras. We started by generating 2 images using Zero-123 at angles $(\theta, \phi) = (10^\circ\pm \delta, 90^\circ)$ and saved all denoising predictions from each timestep. Our goal was to average all pairs of predictions to synthesize a 3-$nd$ view at $(10^\circ, 90^\circ)$ for free. We experimented with varying $\delta\in \{1^\circ, 5^\circ, 10^\circ, 20^\circ\}$, and compared them with the full denoising predictions.

Figure~\ref{fig:free-nv} demonstrates that for angles ($\delta$) up to $5^\circ$, novel views closely match fully generated ones, proving effective for closely positioned cameras. Yet, interpolations between cameras at wider angles yield blurrier images. Additionally, optimal window sizes vary by object; for example, a $\delta=5^\circ$ suits the \texttt{ghost} but not the \texttt{capybara}, indicating that best window size is sample-specific.

Based on these insights, we presents a novel approach: instead of computing the noise prediction for every new camera pose and timestep, we create a memory system to store previously computed features. As such, we can retrieve and reuse these pre-computed features whenever needed. Ideally, this approach could reduces redundant calculations and speeds up the optimization process.

\subsection{Hashing-based Feature Reuse}

In light of our analysis, we developed Hash3D, a solution that incorporates hashing techniques to optimize the SDS. Hash3D is fundamentally designed to minimize the repetitive computational burden typically associated with the diffusion model, effectively trading storage space for accelerated 3D optimization.

At the core of Hash3D is a hash table for storing and retrieving previously extracted features. When Hash3D samples a specific camera pose $\bm c$ and timestep $t$, it first checks the hash table for similar features. If a match is found, it's reused directly in the diffusion model, significantly cutting down on computation. If there's no match in the same hash bucket, the model performs standard inference, and the new feature is added to the hash table for future use.

\noindent\textbf{Grid-based Hashing.} For efficient indexing in our hash table, we use a \emph{grid-based hashing function} with keys composed of camera poses \(\bm c = (\theta,\phi,\rho)\) and timestep \(t\). This function assigns each camera and timestep to a designated grid cell, streamlining data organization and retrieval.

Firstly, we define the size of our grid cells in both the spatial and temporal domains, denoted as \(\Delta \theta, \Delta \phi, \Delta \rho\) and \(\Delta t\) respectively. For each input key \([\theta,\phi,\rho, t]\), the hashing function calculates the indices of the corresponding grid cell. This is achieved by dividing each coordinate by its respective grid size
{\small
\begin{align}
    i = \left\lfloor \frac{\theta}{\Delta \theta} \right\rfloor, j = \left\lfloor \frac{\phi}{\Delta \phi} \right\rfloor, k = \left\lfloor \frac{\rho}{\Delta \rho} \right\rfloor, l = \left\lfloor \frac{t}{\Delta t} \right\rfloor
\end{align}}
Upon obtaining these indices, we combine them into a single hash code that uniquely identifies each bucket in the hash table. The hash function \(\texttt{idx} = (i + N_1 \cdot j +  N_2 \cdot k + N_3 \cdot l)\mod n \) is used, where \(N_1, N_2, N_3\) are large prime numbers~\cite{teschner2003optimized,10.1145/2508363.2508374}, and $n$ denotes the size of the hash table. 

Through this hash function, keys that are close in terms of camera pose and timestep are likely to be hashed to the same bucket. This grid-based approach not only making the data retrieval faster but also maintains the spatial-temporal relationship inherent in the data, which is crucial for our method.

\noindent\textbf{Collision Resolution}. 
When multiple keys are assigned to the same hash value, a collision occurs. We address these collisions using \emph{separate chaining}.
In this context, each hash value \(\texttt{idx}\) is linked to a distinct queue, denoted as \(q_{\texttt{idx}} \). To ensure the queue reflects the most recent data and remains manageable in size, it is limited to a maximum length $Q=3$. When this limit is reached, the oldest elements is removed to accommodate the new entry, ensuring the queue stays relevant to the evolving 3D representation.



\begin{figure*}[!tb]
    \centering
    \includegraphics[width=\linewidth]{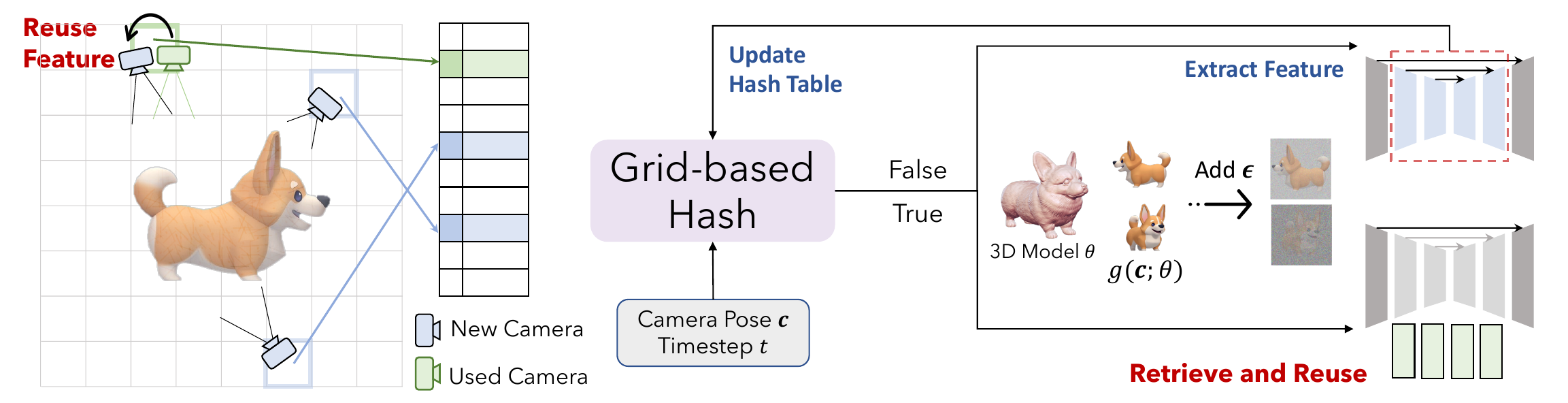}
    \caption{Overall pipeline of our Hash3D. Given the sampled camera and time-step, we retrieve the intermediate diffusion feature from hash table. If no matching found, it performs a standard inference and stores the new feature in the hash table; otherwise, if a feature from a close-up view already exists, it is reused without re-calculation.}
    \label{fig:enter-label}
    \vspace{-5mm}
\end{figure*}

\noindent\textbf{Feature Retrieval and Update}.
Once the hash value \texttt{idx} is determined, we either retrieve existing data from the hash table or update it with new features. We set a hash probability $0 <\eta < 1$ to make sure the balanced behavior between retrieval and update. In other words, with probability $\eta$, we retrieve the feature; otherwise, it performs hash table updates. 

For feature updates, following prior work~\cite{ma2023deepcache}, we extract the feature $\mathbf{v}_{l-1}^{(U)}$, which is the input of last up-sampling layer in the U-net. Once extracted, we compute the hash code \(\texttt{idx}\) and append the data to the corresponding queue $q_{\texttt{idx}}$. The stored data includes input noisy latent $\bm x$, camera pose $\bm c$, timestep $t$, and extracted diffusion features $\mathbf{v}_{l-1}^{(U)}$.

For feature retrieval, we aggregate data from \( q_{\texttt{idx}} \) through weighted averaging. This method considers the distance of each noisy input $\bm x_i$ from the current query point $\bm x$. The weighted average \( \mathbf{v} \) for a given index is calculated as follows:
{\small\begin{align}
    \mathbf{v} = \sum_{i=1}^{|q_{\texttt{idx}}|} W_i \mathbf{v}_i, \text{ where } W_i = \frac{e^{(-||\bm x- \bm x_i||_2^2)}}{\sum_{i=1}^{|q_{\texttt{idx}}|} e^{(-||\bm x- \bm x_i||_2^2)}}
\end{align}}
Here, \( W_i \) is the weight assigned to \( \mathbf{v}_i \) based on its distance from the query point, and \( |q_{\texttt{idx}}| \) is the current length of the queue. An empty queue $|q_{\texttt{idx}}|$ indicates unsuccessful retrieval, necessitating feature update.

\subsection{Adaptive Grid Hashing}

In grid-based hashing, the selection of an appropriate grid size \(\Delta \theta, \Delta \phi, \Delta \rho, \Delta t\) — plays a pivotal role. As illustrated in Section~\ref{sec:redundency}, we see three insights related to grid size. First, feature similarity is only maintained at a median grid size; overly large grids tend to produce artifacts in generated views. Second, it is suggested that ideal grid size differs across various objects. Third, even for a single object, optimal grid sizes vary for different views and time steps, indicating the necessity for adaptive grid sizing to ensure optimal hashing performance.

\noindent\textbf{Learning to Adjust the Grid Size.} To address these challenges, we propose to dynamically adjusting grid sizes. The objective is to maximize the average cosine similarity $\text{cos}(\cdot,\cdot)$ among features within each grid. In other words, only if the feature is similar enough, we can reuse it. Such problem is formulated as
{\small
\begin{equation}
    \max_{\Delta \theta, \Delta \phi, \Delta \rho, \Delta t} \frac{1}{|q_{\texttt{idx}}| } \sum_{i,j}^{|q_{\texttt{idx}}|} \text{cos}(\mathbf{v}_j,\mathbf{v}_i), \quad s.t. |q_{\texttt{idx}}| > 0 \quad [\text{Non-empty}]
\end{equation}}

Given our hashing function is \emph{non-differentiale}, we employ a brute-force approach. Namely, we evaluate $M$ predetermined potential grid sizes, each corresponding to a distinct hash table, and only use best one.

For each input \([\theta,\phi,\rho, t]\), we calculate the hash code $\{{\texttt{idx}}^{(m)}\}_{m=1}^M$ for $M$ times, and indexing in each bucket. 
 Feature vectors are updated accordingly, with new elements being appended to their respective bucket. We calculate the cosine similarity between the new and existing elements in the bucket, maintaining a running average $s_{\texttt{idx}^{(n)}}$ of these similarities 
{\small
\begin{equation}
    s_{\texttt{idx}^{(m)}} \leftarrow \gamma s_{\texttt{idx}^{(m)}} + (1-\gamma) \frac{1}{|q_{\texttt{idx}^{(m)}}|} \sum_{i=1}^{|q_{\texttt{idx}^{(m)}}|} \text{cos}(\mathbf{v}_{new},\mathbf{v}_i)
\end{equation}}

During retrieval, we hash across all $M$ grid sizes but only consider the grid with the highest average similarity for feature extraction.

\noindent\textbf{Computational and Memory Efficiency.} Despite employing a brute-force approach that involves hashing $M$ times for each input, our method maintains computational efficiency due to the low cost of hashing. It also maintains memory efficiency, as hash tables store only references to data. To prioritize speed, we deliberately avoid using neural networks for hashing function learning.

\section{Experiment}
In this section, we assess the effectiveness of our HS by integrating it with various 3D generative models, encompassing both image-to-3D and text-to-3D tasks.

\subsection{Experimental Setup}

\noindent\textbf{Baselines.} To validate the versatility of our method, we conducted extensive tests across a wide range of baseline text-to-3D and image-to-3D methods.
\begin{itemize}
    \item \textbf{Image-to-3D.}  Our approach builds upon techniques such as Zero-123+SDS ~\cite{Liu_2023_ICCV}, DreamGaussian~\cite{tang2023dreamgaussian} and Magic123~\cite{Magic123}. For Zero-123+SDS, we have incorporated Instant-NGP~\cite{mueller2022instant} and Gaussian Splatting~\cite{kerbl3Dgaussians} as its representation. We call these two variants Zero-123~(NeRF) and Zero-123~(GS).
    \item \textbf{Text-to-3D.} Our tests also covered a range of methods, such as Dreamfusion~\cite{poole2023dreamfusion}, Fantasia3D~\cite{Chen_2023_ICCV}, Latent-NeRF~\cite{10205242}, Magic3D~\cite{lin2023magic3d}, and GaussianDreamer~\cite{yi2023gaussiandreamer}.
\end{itemize}
For DreamGaussian and GaussianDreamer, we implement Hash3D on top of the official code. And for other methods, we use the reproduction from \texttt{threestudio}\footnote{https://github.com/threestudio-project/threestudio}.

\noindent\textbf{Implementation Details.} We stick to the same hyper-parameter setup within their original implementations of these methods. For text-to-3D, we use the \texttt{stable-diffusion-2-1}\footnote{https://huggingface.co/stabilityai/stable-diffusion-2-1} as our 2D diffusion model. For image-to-3D, we employ the \texttt{stable-zero123}\footnote{https://huggingface.co/stabilityai/stable-zero123}. We use a default hash probability setting of $\eta=0.1$. We use $M=3$ sets of grid sizes, with $\Delta \theta, \Delta \phi, \Delta t\in \{10, 20,30\}$ and $\Delta \rho \in\{0.1, 0.15, 0.2\}$. We verify this hyper-parameter setup in the ablation study. 

\noindent\textbf{Dataset and Evaluation Metrics.} To assess our method, we focus on evaluating the computational cost and visual quality achieved by implementing Hash3D.
\begin{itemize}[noitemsep,topsep=0pt]
    \item \textbf{Image-to-3D.}  For the image-to-3D experiments, we leverage the Google Scanned Objects (GSO) dataset~\cite{downs2022google} for evaluation~\cite{liu2024one,liu2023zero}. We focused on evaluating novel view synthesis (NVS) performance using established metrics such as PSNR, SSIM~\cite{wang2004image}, and LPIPS~\cite{zhang2018unreasonable}. We selected 30 objects from the dataset. For each object, we generated a 256$\times$256 input image for 3D reconstruction. We then rendered 16 different views at a 30-degree elevation, varying azimuth angles, to compare the reconstructed models with their ground-truth. To ensure semantic consistency, we also calculated CLIP-similarity scores between the rendered views and the original input images. 
    \item \textbf{Text-to-3D.} We generated 3D models from 50 different prompts, selected based on a prior study. To evaluate our methods, we focused on two primary metrics: mean$\pm$std CLIP-similarity~\cite{pmlr-v139-radford21a,qian2023magic123,liu2023one2345} and the average generation time for each method. For assessing CLIP-similarity, we calculated the similarity between the input text prompt and 8 uniformly rendered views at elevation $\phi=0^{\circ}$ and azimuth $\theta=[0^{\circ}, 45^{\circ}, 90^{\circ}, 135^{\circ}, 180^{\circ}, 225^{\circ}, 270^{\circ}, 315^{\circ}]$. Additionally, we recorded and reported the generation time for each run.
    \item \textbf{User Study.} To evaluate the visual quality of generated 3D objects, we carried out a study involving 44 participants. They were shown 12 videos of 3D renderings, created using two methods: Zero-123~(NeRF) for images-to-3D, and Gaussian-Dreamer for text-to-3D. These renderings were made both with and without Hash3D. Participants were asked to rate the visual quality of each pair of renderings, distributing a total of 100 points between the two in each pair to indicate their perceived quality difference.
    \item \textbf{Computational Cost.} We report the running time for each experiment using a single RTX A5000. Besides, we report MACs in the tables. Given that feature retrieval is stochastic — implying that retrieval of features is not guaranteed with attempt in empty bucket — we provide the theoretical average MACs across all steps, pretending that all retrieval succeeded.
\end{itemize}

\begin{figure}[t]
    \centering
    \includegraphics[width=\linewidth]{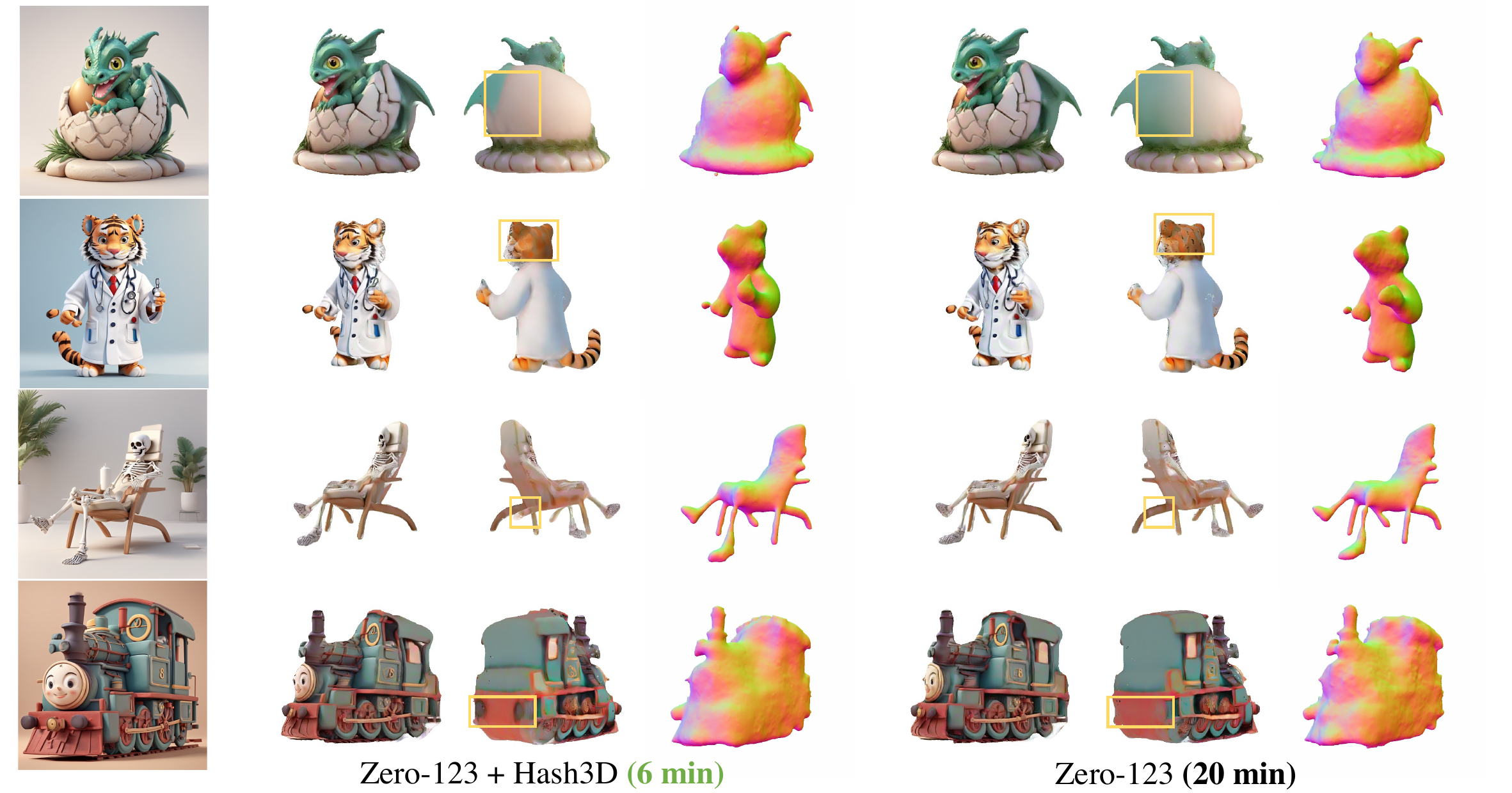}
    \caption{Qualitative Results using Hash3D along with Zero123 for image-to-3D generation. We mark the visual dissimilarity in \textcolor{amber}{yellow}.}
    \vspace{-3mm}
    \label{fig:Zero123}
\end{figure}

\subsection{3D Generation Results}

\noindent\textbf{Image-to-3D Qualitative Results.} Figure~\ref{fig:Zero123} demonstrates the outcomes of incorporating Hash3D into the Zero-123 framework to generate 3D objects. This integration not only preserves visual quality and consistency across views but also markedly decreases the processing time. In specific instances, Hash3D outperforms the baseline, as evidenced by the enhanced clarity of the dragon wings' boundaries in row 1 and the more distinct taillights of the train in row 4. A similar level of visual fidelity is observed in Figure~\ref{fig:head}, where Hash3D is applied in conjunction with DreamGaussian, indicating that the integration effectively maintains quality while improving efficiency.

\noindent\textbf{Image-to-3D Quantitative Results.} For a detailed numerical analysis, refer to Table~\ref{tab:image-to-3D}, which outlines the novel view synthesis performance, CLIP scores, running times for on top of all 4 baseline methods.  Notably, For DreamGaussian and Zero-123(NeRF), we speed up the running time by $4\times$ and $3\times$ respectively. This reduction in running times is mainly due to the efficient feature retrieval and reuse mechanism employed by Hash3D. Additionally, our approach not only speeds up the process but also slightly improves performance. We believe this enhancement stems from the sharing of common features across different camera views, which reduces the inconsistencies found in independently sampled noise predictions, resulting in the smoother generation of 3D models.

\begin{table}[!t]
\renewcommand{\arraystretch}{1.2}
    \centering
    \scriptsize
    \caption{Speed and performance comparison when integrated image-to-3D models with Hash3D. We report the original running time in their paper. }
    \label{tab:image-to-3D}
    \vspace{-2mm}
    \begin{tabular}{l|r|c|c|c|c|c|c|c}
    \toprule
        \textbf{Method} & \textbf{Time$\downarrow$} & \textbf{Speed$\uparrow$} & \textbf{MACs$\downarrow$} &  \textbf{PSNR$\uparrow$} & \textbf{SSIM$\uparrow$} & \textbf{LPIPS$\downarrow$} &&  \textbf{CLIP-G/14$\uparrow$} \\
        \midrule
        DreamGaussian & 2m & - & 168.78G &  16.202{\tiny$\pm$2.501} &	0.772{\tiny$\pm$0.102}&	0.225{\tiny$\pm$0.111} && 0.693{\tiny$\pm$0.105}\\
        + \textbf{Hash3D} &  \textbf{\cellcolor{green!25}30s} & \textbf{\cellcolor{green!25}4.0$\times$} & 154.76G & \textbf{16.356}{\tiny$\pm$2.533}	&\textbf{0.776}{\tiny$\pm$0.103} & \textbf{0.223}{\tiny$\pm$0.113} &&  \textbf{0.694}{\tiny$\pm$0.104}\\
        \midrule
        Zero-123(NeRF) &20m & - & 168.78G & 17.773{\tiny$\pm$3.074} &0.787{\tiny$\pm$0.101}	& 0.198{\tiny$\pm$0.097} & & 0.662{\tiny$\pm$0.0107}  	\\
        + \textbf{Hash3D} & \textbf{\cellcolor{green!25}7m} & \textbf{\cellcolor{green!25}3.3$\times$}  & 154.76G &\textbf{17.961}{\tiny$\pm$3.034}	& \textbf{0.789}{\tiny$\pm$0.095}	& \textbf{0.196}{\tiny$\pm$0.0971} && \textbf{0.665}{\tiny$\pm$0.104}\\
        \midrule 
        Zero-123(GS) &6m & - & 168.78G & 18.409{\tiny$\pm$2.615}  &0.789{\tiny$\pm$0.100}	& \textbf{0.204}{\tiny$\pm$0.101} & & 0.643{\tiny$\pm$0.105} 	\\
        + \textbf{Hash3D} & \textbf{\cellcolor{green!25}3m} &\textbf{\cellcolor{green!25}2.0$\times$}& 154.76G&   \textbf{18.616}{\tiny$\pm$2.898}	& \textbf{0.793}{\tiny$\pm$0.099} & \textbf{0.204}{\tiny$\pm$0.099} && \textbf{0.632}{\tiny$\pm$0.106}\\
                \midrule
        Magic123 & 120m & - & 847.38G &\textbf{18.718}{\tiny$\pm$2.446}
        &\textbf{0.803}{\tiny$\pm$0.093} & \textbf{0.169}{\tiny$\pm$0.092} &&\textbf{0.718}{\tiny$\pm$0.099}\\
        + \textbf{Hash3D} & \textbf{\cellcolor{green!25}90m} & \textbf{\cellcolor{green!25}1.3$\times$}  & 776.97G&  18.631{\tiny$\pm$2.726} & \textbf{0.803}{\tiny$\pm$0.091}& 0.174{\tiny$\pm$0.093} && 0.715{\tiny$\pm$0.107}  \\
        \bottomrule
    \end{tabular}
    \vspace{-3mm}
\end{table}
\begin{table}[t]
\renewcommand{\arraystretch}{1.2}
    \centering
     \scriptsize
    \caption{Speed and performance comparison between various text-to-3D baseline when integrated with Hash3D.}
    \vspace{-2mm}
    \label{tab:text23d}
    \setlength{\tabcolsep}{3pt}
    \begin{tabular}{l|r|c|c|c|c|c}
    \toprule
        \textbf{Method} & \textbf{Time$\downarrow$} & \textbf{Speed$\uparrow$} & \textbf{MACs$\downarrow$} & \textbf{CLIP-G/14$\uparrow$} & \textbf{CLIP-L/14$\uparrow$} & \textbf{CLIP-B/32$\uparrow$}\\
        \midrule
        Dreamfusion & 1h 00m &- &678.60G & 0.407{\tiny$\pm$ 0.088} & \textbf{0.267}{\tiny$\pm$0.058} & \textbf{0.314} {\tiny $\pm$0.049}\\
        + \textbf{Hash3D} & \textbf{\cellcolor{green!25}40m} & \textbf{\cellcolor{green!25}1.5$\times$} &622.21G &\textbf{0.411}{\tiny$\pm$0.070} & 0.266{\tiny$\pm$ 0.050}&0.312{\tiny$\pm$0.044}\\
        \hline
        Latent-NeRF & 30m &-& 678.60G& \textbf{0.406}{\tiny$\pm$0.033} & 0.254{\tiny$\pm$0.039} & 0.306{\tiny$\pm$0.037}\\
        + \textbf{Hash3D} & \textbf{\cellcolor{green!25}17m} & \textbf{\cellcolor{green!25}1.8$\times$} &622.21G & \textbf{0.406}{\tiny$\pm$0.038} & \textbf{0.258}{\tiny$\pm$0.045} &\textbf{0.305}{\tiny$\pm$0.038}\\
        \hline
        SDS+GS & 1h 18m & -& 678.60G & \textbf{0.413}{\tiny$\pm$0.048} & \textbf{0.263}{\tiny$\pm$0.034} & \textbf{0.313}{\tiny$\pm$0.036}\\
        + \textbf{Hash3D} & \textbf{\cellcolor{green!25}40m} & \textbf{\cellcolor{green!25}1.9$\times$}&622.21G& 0.402{\tiny$\pm$0.062} & 0.252{\tiny$\pm$0.041} & 0.306{\tiny$\pm$0.036}\\
        \hline
        Magic3D & 1h 30m &-& 678.60G & \textbf{0.399}{\tiny$\pm$0.012} & \textbf{0.257}{\tiny$\pm$0.064} & \textbf{0.303}{\tiny$\pm$0.059}\\
        
        + \textbf{Hash3D} & \textbf{\cellcolor{green!25}1h}& \textbf{\cellcolor{green!25}1.5$\times$}&622.21G & 0.393{\tiny$\pm$0.011} &0.250{\tiny$\pm$0.054} & 0.304{\tiny$\pm$0.052}\\
        \bottomrule
        GaussianDreamer & 15m &-& 678.60G &  0.412{\tiny$\pm$0.049}& 0.267{\tiny$\pm$0.035} & 0.312{\tiny$\pm$0.038}\\
        
        + \textbf{Hash3D} &  \textbf{\cellcolor{green!25}10m} & \textbf{\cellcolor{green!25}1.5$\times$}&622.21G &\textbf{0.416}{\tiny$\pm$0.057} & \textbf{0.271}{\tiny$\pm$0.036}  & \textbf{0.312}{\tiny$\pm$0.037}\\
        \bottomrule
    \end{tabular}
     \vspace{-3mm}
\end{table}

\begin{figure*}[t]
    \centering
    \includegraphics[width=\linewidth]{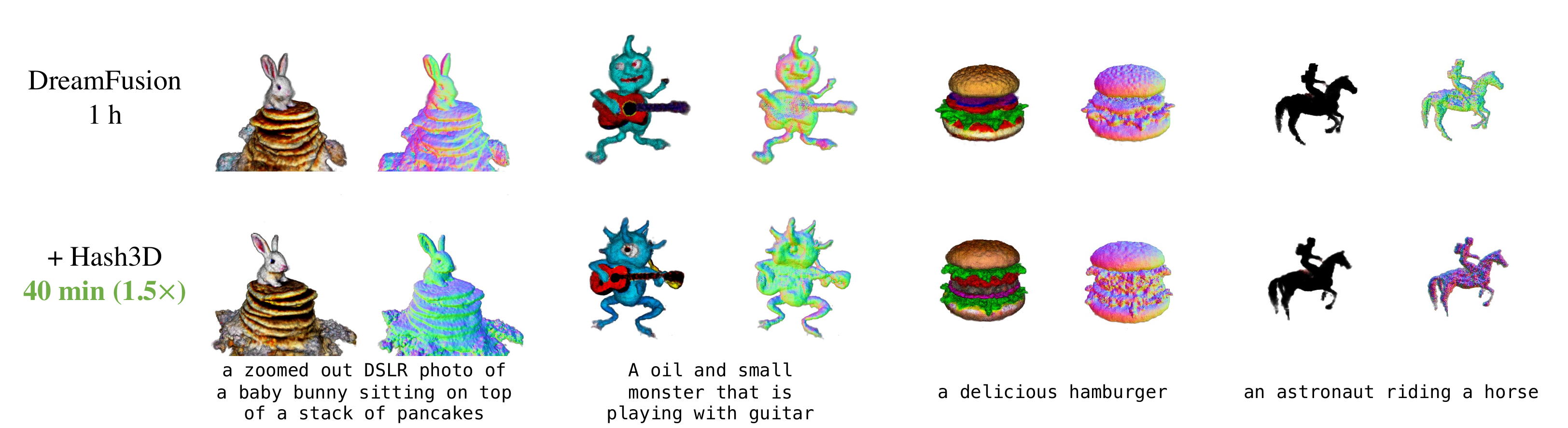}
    \includegraphics[width=\linewidth]{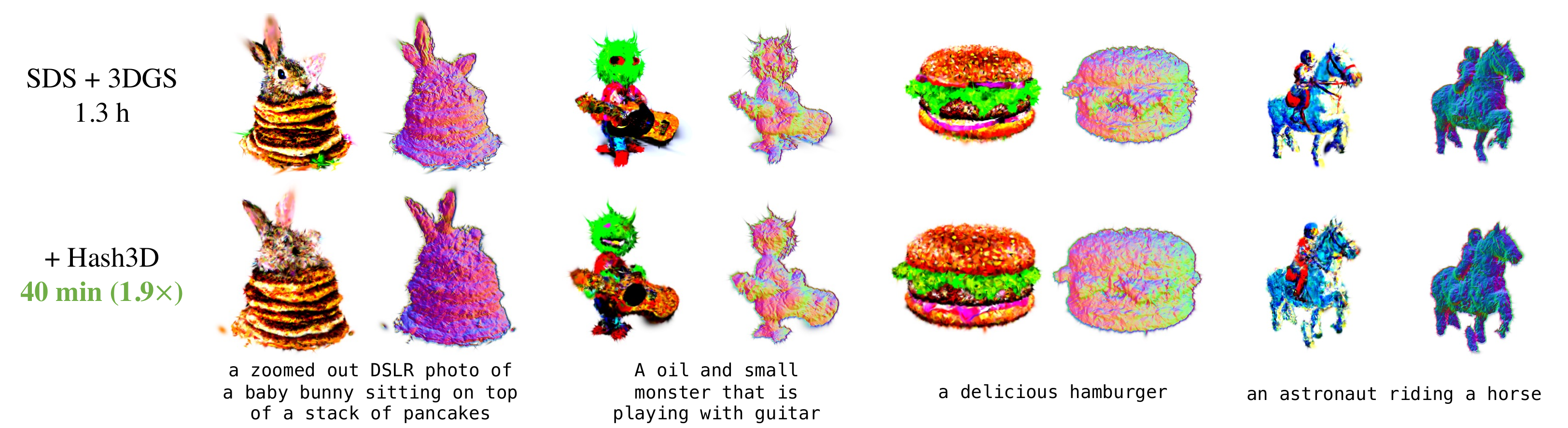}
    \includegraphics[width=\linewidth]{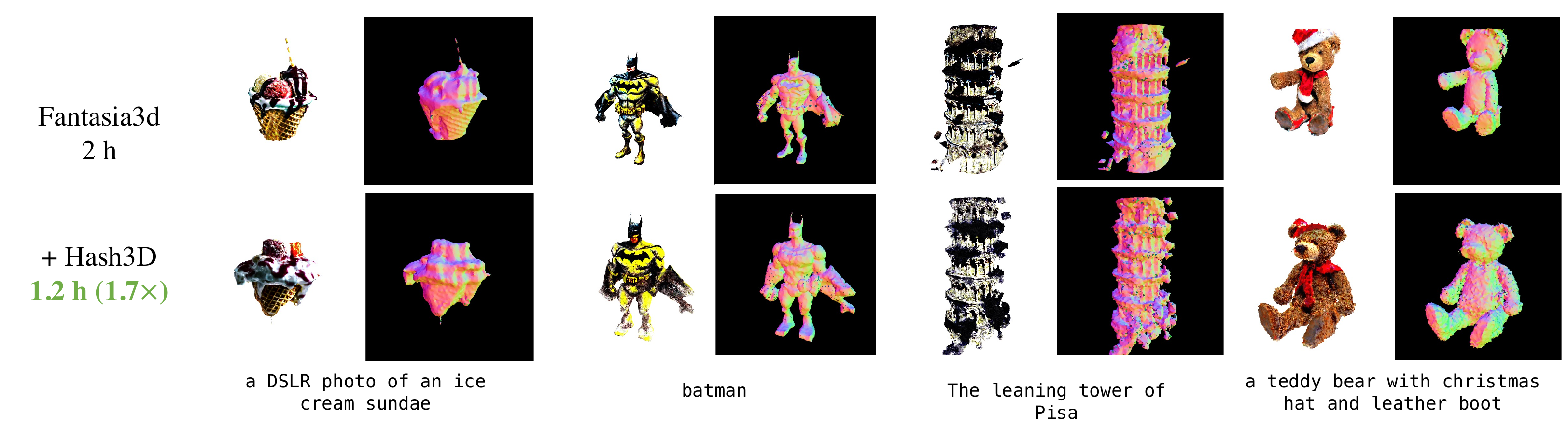}
    \caption{Visual comparison for text-to-3D task, when applying Hash3D to DreamFusion~\cite{poole2023dreamfusion}, SDS+GS and Fantasia3D~\cite{Chen_2023_ICCV}.}
    \label{fig:text23d}
\end{figure*}


\noindent\textbf{Text-to-3D Qualitative Results.} In Figure~\ref{fig:text23d}, we present the results generated by our method, comparing Hash3D with DreamFusion~\cite{poole2023dreamfusion}, SDS+GS, and Fantasia3D~\cite{Chen_2023_ICCV}. The comparison demonstrates that Hash3D maintains comparable visual quality to these established methods.

\noindent\textbf{Text-to-3D Quantitative Results.} Table~\ref{tab:text23d} provides a detailed quantitative evaluation of Hash3D. Across various methods, Hash3D markedly decreases processing times, showcasing its adaptability in speeding up 3D generation. Significantly, this reduction in time comes with minimal impact on the CLIP score, effectively maintaining visual quality. Notably, with certain methods such as GaussianDreamer, Hash3D goes beyond maintaining quality; it subtly improves visual fidelity. This improvement suggests that Hash3D's approach, which considers the relationship between nearby camera views, has the potential to enhance existing text-to-3D generation processes.

\begin{wrapfigure}{r}{0.6\textwidth}
\vspace{-13mm}
  \begin{center}
  \centering
    \includegraphics[width=\linewidth]{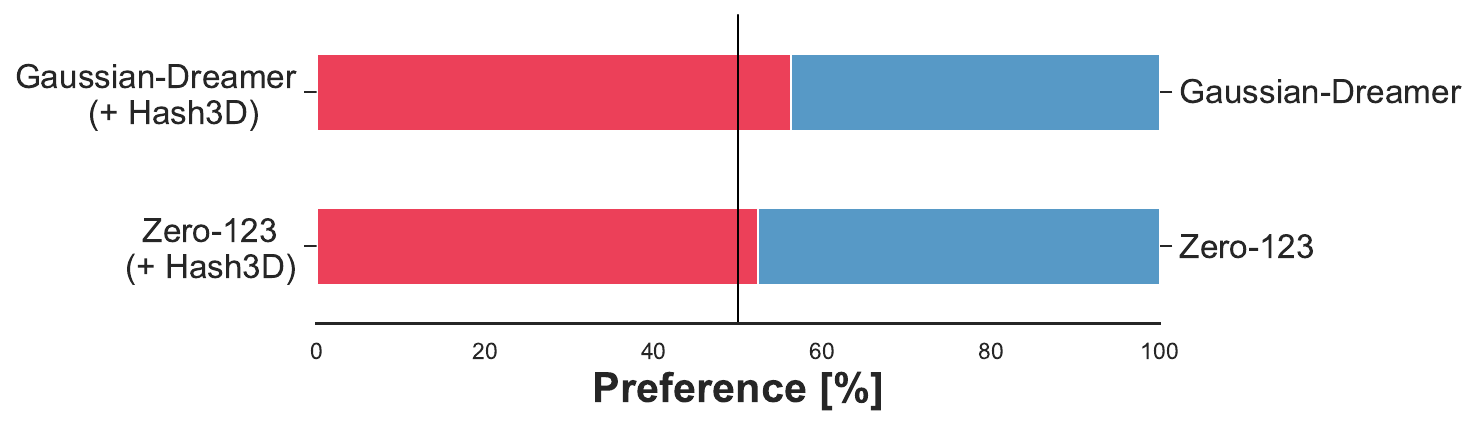}
  \end{center}
  \vspace{-7mm}
  \caption{User preference study for Hash3D.}
    \label{fig:user_study}
    \vspace{-9mm}
\end{wrapfigure} 
\noindent\textbf{User preference study.} As shown in Figure~\ref{fig:user_study}, Hash3D received an average preference score of 52.33/100 and 56.29/100 when compared to Zero-123 (NeRF) and Gaussian-Dreamer. These scores are consistent with previous results, indicating that Hash3D slightly enhances the visual quality of the generated objects.


\subsection{Ablation Study and Analysis}

In this section, we study several key components in our Hash3D framework.

\begin{figure}[t]
\centering
\begin{minipage}{.42\textwidth}
  \centering
    \includegraphics[width=\linewidth]{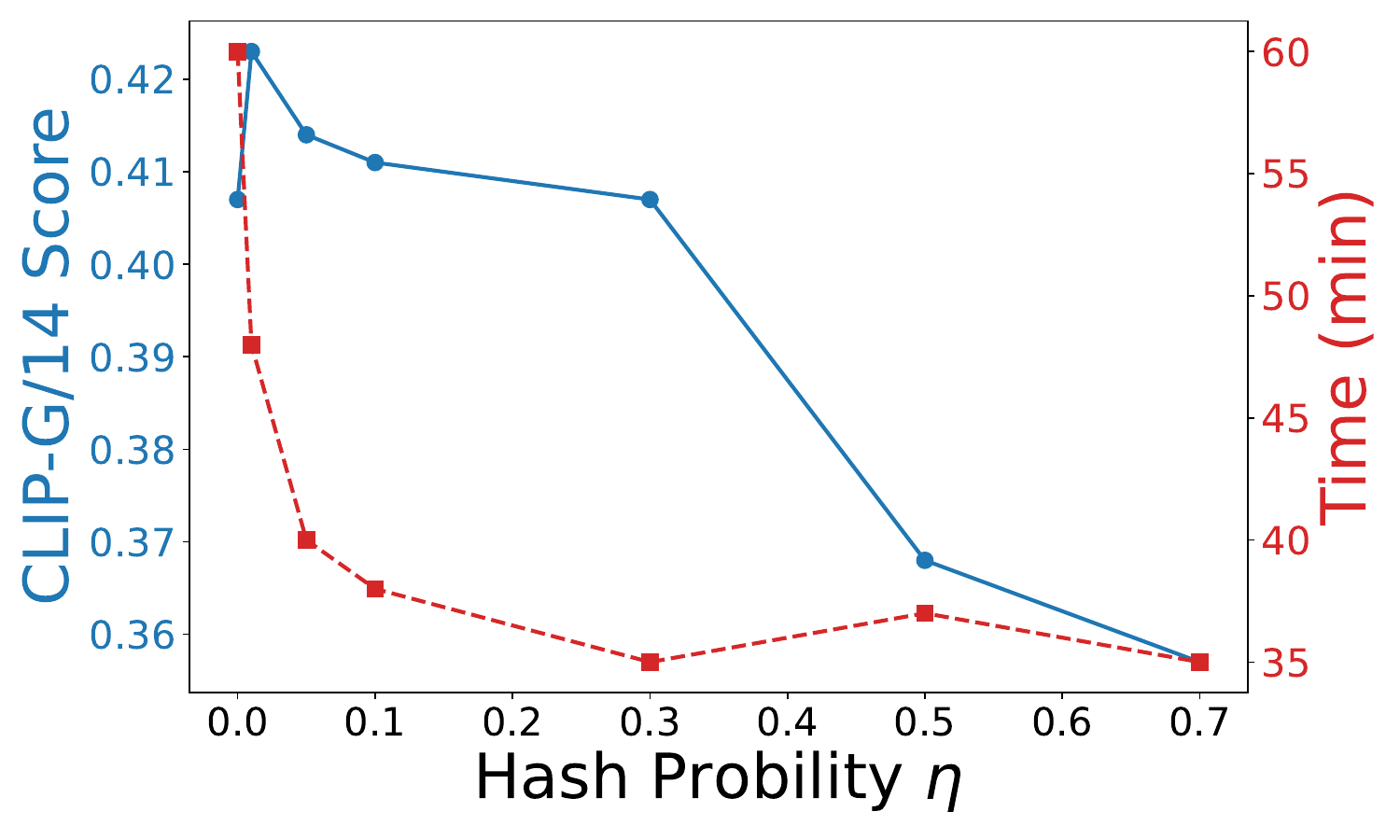}
    \vspace{-8mm}
    \caption{Ablation study with different hash probability $\eta$.}
    \label{fig:ablation_eta}
    \vspace{-3mm}
\end{minipage}%
\hfill
\begin{minipage}{.55\textwidth}
\renewcommand{\arraystretch}{1.6}
\scriptsize
  \centering
      \begin{tabular}{l|c|c}
    \toprule
        \textbf{Method} & \textbf{Time}  & \textbf{CLIP-G/14} \\
    \midrule
        Zero-123~(NeRF) + Hash3D w/n & 6 min  & 0.631{\tiny$\pm$0.090} \\
        Zero-123~(NeRF) + Hash3D & 7 min & \textbf{0.665}{\tiny$\pm$0.104 }\\
    \midrule
        Zero-123~(GS) + Hash3D w/n   & 3 min  & 0.622$\pm$0.083 \\
        Zero-123~(GS) + Hash3D    & 3 min & \textbf{0.632}{\tiny$\pm$1.06} \\
    \bottomrule
    \end{tabular}
    \caption{Comparison between Hashing Features \emph{vs.} Hashing Noise, applied to Zero-123.}
    \label{tab:Hash3D_ablation}

\end{minipage}
\end{figure}

\noindent\textbf{Ablation 1: Hashing Features \emph{vs.} Hashing Noise.} Our Hash3D involves hashing intermediate features in the diffusion U-Net. Alternatively, we explored hashing the predicted \emph{noise} estimation directly, leading to the development of a variant named Hash3D with noise (Hash3D w/n). This variant hashes and reuses the predicted score function directly. We applied this approach to the image-to-3D task using Zero123, and the results are detailed in Table~\ref{tab:Hash3D_ablation}. Interestingly, while Hash3D w/n demonstrates a reduction in processing time, it yields considerably poorer results in terms of CLIP scores. This outcome underscores the effectiveness of our initial choice to hash features rather than noise predictions.



\noindent\textbf{Ablation 2: Influence of Hash Probability $\eta$.} A crucial factor in our Hash3D  is the feature retrieval probability $\eta$. To understand its impact, we conducted an ablation experiment with Dreamfusion, testing various $\eta$ values $\{0.01, 0.05, 0.1, 0.3, 0.5, 0.7\}$.

\begin{figure*}[t]
    \centering
    \includegraphics[width=\linewidth]{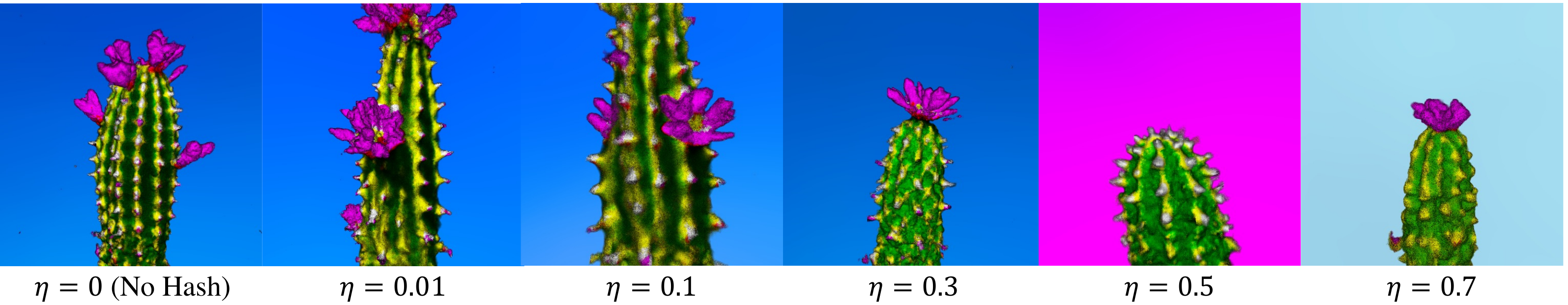}
    \caption{Quantitative ablation study with different hash probability $\eta$}
    \label{fig:eta}
   
     \vspace{-2mm}
\end{figure*}

The relationship between CLIP score, time, and different $\eta$ values is depicted in Figure~\ref{fig:ablation_eta}. We observed that running time steadily decrease across all values. Interestingly, with smaller $\eta$ values (less than 0.3), Hash3D even improved the visual quality of the generated 3D models. We speculate this improvement results from the enhanced smoothness in predicted noises across different views, attributable to feature sharing via a grid-based hash table. However, when $\eta>0.3$, there was negligible impact on running time reduction. Figure~\ref{fig:eta} showcases the same trend in terms of visual quality. A moderately small $\eta$ effectively balances performance and efficiency. Consequently, we opted for $\eta=0.1$ for the experiments presented in our main paper.

\noindent\textbf{Ablation 3: Adaptive Grid Size.} In this study, we introduce a dynamic adjustment of the grid size for hashing, tailored to each individual sample. This adaptive approach, termed AdaptGrid, is evaluated against a baseline method that employs a constant grid size, within the context of Dreamfusion. As illustrated in Table~\ref{tab:grid_size}, the AdaptGrid strategy surpasses the performance of the constant grid size method. Larger grid sizes tend to compromise the visual quality of generated 3D objects. Conversely, while smaller grid sizes preserve performance to a greater extent, they significantly reduce the likelihood of matching nearby features, resulting in increased computation time.

\begin{table}[h]
    \centering
    \vspace{-2mm}
    \scriptsize
    \begin{tabular}{l|c|c|c|>{\columncolor{green!25}}c}
    \toprule
        $\Delta \theta, \Delta \phi, \Delta \rho, \Delta t$ & (10, 10, 0.1, 10) & (20, 20, 0.15, 20)  & (30, 30, 0.2, 30) & AdaptGrid (Ours)\\
        \midrule
        \textbf{CLIP-G/14}$\uparrow$ & 0.408{\tiny$\pm$0.033} & 0.345{\tiny$\pm$0.055} & 0.287{\tiny$\pm$0.078}&\textbf{0.411}{\tiny$\pm$0.070}\\
        \textbf{Time}$\downarrow$ & 48m & 38m & 32m& 40m \\
        \bottomrule
    \end{tabular}
    \caption{Ablation study on the Adaptive \emph{v.s.} Constant Grid Size.}
    \label{tab:grid_size}
    \vspace{-5mm}
\end{table}

\section{Conclusion}

In this paper, we present Hash3D, a training-free technique that improves the efficiency of diffusion-based 3D generative modeling. Hash3D utilizes adaptive grid-based hashing to efficiently retrieve and reuse features from adjacent camera poses, to minimize redundant computations. As a result, Hash3D not only speeds up 3D model generation by $1.3 \sim 4 \times$ without the need for additional training, but it also improves the smoothness and consistency of the generated 3D models.

%
%
\bibliographystyle{splncs04}
\bibliography{main}
\end{document}